# Hybrid Bayesian Networks with Linear Deterministic Variables

**Barry R. Cobb**
Department of Economics and Business
Virginia Military Institute
Lexington, VA 24450
*cobbbr@vmi.edu*

**Prakash P. Shenoy**
University of Kansas School of Business
1300 Sunnyside Ave., Summerfield Hall
Lawrence, KS 66045-7585
*pshenoy@ku.edu*

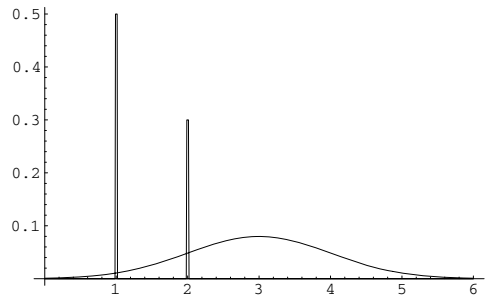

Figure 1: The Mixed Distribution for $X$.

## Abstract

When a hybrid Bayesian network has conditionally deterministic variables with continuous parents, the joint density function for the continuous variables does not exist. Conditional linear Gaussian distributions can handle such cases when the continuous variables have a multi-variate normal distribution and the discrete variables do not have continuous parents. In this paper, operations required for performing inference with conditionally deterministic variables in hybrid Bayesian networks are developed. These methods allow inference in networks with deterministic variables where continuous variables may be non-Gaussian, and their density functions can be approximated by mixtures of truncated exponentials. There are no constraints on the placement of continuous and discrete nodes in the network.

Consider a real-valued random variable $X$ with a mixed distribution where $P(X = 1) = 0.5$, $P(X = 2) = 0.3$, and which has probability density $0.2 \cdot \phi(x)$, where $\phi(x)$ is a normal probability density function (PDF) with mean 3 and variance 1, i.e. $N(3, 1)$. We can represent this mixed distribution by the mixed potential $\zeta$ for $X$ as follows: $\zeta(X = 1) = (0.5, \iota)$; $\zeta(X = 2) = (0.3, \iota)$; and $\zeta(x) = (1, 0.2 \cdot \phi(x))$ for $x \neq 1, 2$.

## 1 INTRODUCTION

An important class of Bayesian networks with discrete and continuous variables are those that have conditionally deterministic variables (a variable that is a deterministic function of its parents). Conditional linear Gaussian (CLG) models [Lauritzen and Jensen 2001] can handle such cases when continuous variables have a multi-variate Gaussian distribution and discrete nodes do not have continuous parents. In models with non-Gaussian distributions and deterministic variables, Monte Carlo methods may be required to obtain an approximate solution. General purpose solution algorithms, e.g., the Shenoy-Shafer architecture, have not been adapted to such models, primarily because the joint density for the variables in models with deterministic variables does not exist and these methods involve propagation of probability densities.

The first component of the value of a mixed potential is called a mass part and the second component is called a density part. The symbol $\iota$ represents a vacuous density or an absence of a probability density. When we have a value in the mass part with an absence of density in the density part, the value in the mass part represents probability mass. When we have values in both parts, the product of the two values represents probability density. In this case, the value in the mass part can be interpreted as a "weight" for the density value in the density part. For convenience, henceforth we will omit the qualification "for $x \neq 1, 2$." We can verify $\zeta$ is a probability distribution for $X$ by calculating: $0.5 + 0.3 + \int_{\Omega_X} 0.2 \cdot \phi(x)\, dx = 1$. This mixed distribution is shown graphically in Figure 1.

Mixed distributions result from the computation of marginal distributions for variables in hybrid Bayesian networks, such as the network shown in Figure 2. In this example, $Y$ is a discrete variable with state space $\Omega_Y = \{1, 2, 3\}$, $Z$ is a real-valued variable, and $X$ is a conditionally deterministic variable whose conditional distribution is described as follows: $X \mid \{y, z\} = y$ if $y = 1, 2$ and $X \mid \{y, z\} = z$ if $y = 3$.

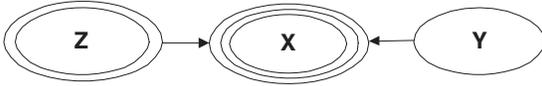

Figure 2: A Bayesian Network Representation of the Mixed Distribution for $X$.

Potentials of the type described above are used in this paper to represent distributions of variables in hybrid Bayesian networks where continuous variables are (possibly) linear deterministic functions of their continuous parents. This paper develops inference methods using mixed potentials that allow inference in hybrid Bayesian networks without placing limitations on the types of continuous distributions allowed and without limitations on the placement of continuous and discrete variables. We use the Shenoy-Shafer architecture to compute marginals and we can use any join tree to do the propagation.

This paper expands on the notation and operations described by Cobb and Shenoy [2004a]. The operations described in Cobb and Shenoy [2004a] were complete only for those hybrid Bayesian networks in which the marginal distributions were either discrete (for discrete variables) or continuous (for continuous variables). Here we define notation and additional operations to allow for the possibility of mixed distribution marginals for continuous variables.

The outline of the remainder of the paper is as follows. Section 2 introduces notation and definitions used throughout the paper. Section 3 defines operations on mixed potentials required for propagation in hybrid Bayesian networks. Section 4 contains a solved example using the operations defined in the paper. Section 5 summarizes the paper. This paper is extracted from a larger, unpublished working paper [Cobb and Shenoy 2005a].

## 2 NOTATION AND DEFINITIONS

### 2.1 Notation

Random variables in a hybrid Bayesian network will be denoted by capital letters, e.g., $A, B, C$. Sets of variables will be denoted by boldface capital letters, $\mathbf{Y}$ if all variables are discrete, $\mathbf{Z}$ if all variables are continuous, or $\mathbf{X}$ if some of the components are discrete and some are continuous. If $\mathbf{X}$ is a set of variables, $\mathbf{x}$ is a configuration of specific states of those variables. The discrete, continuous, or mixed state space of $\mathbf{X}$ is denoted by $\Omega_\mathbf{X}$. Mixed potentials are denoted by lower-case greek letters, e.g., $\alpha$, $\beta$, $\gamma$, possibly with subscripts as an index.

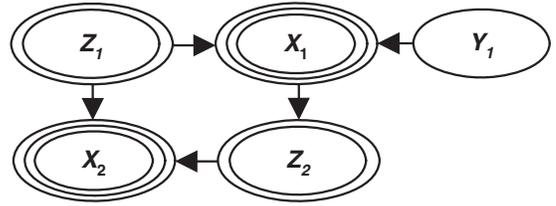

Figure 3: The Bayesian network for Example 1.

In graphical representations, continuous variables in hybrid Bayesian networks are represented by double-border ovals, whereas discrete variables are represented by single-border ovals. Variables that are conditionally deterministic functions of their parents are represented by triple-border ovals. A variable is discrete if its state space is countable, and continuous otherwise. The variable $X$ described in Section 1 is continuous because its state space is the set of real numbers.

**Example 1** Consider the Bayesian network in Figure 3. In this model, $Y_1$ is a discrete variable which can take on values $Y_1 = 0$ or $Y_1 = 1$. The variables $Z_1$ and $Z_2$ are continuous variables whose state spaces are the set of real numbers. The variable $X_1$ is conditionally deterministic given $\{y_1, z_1\}$. The variable $X_2$ is conditionally deterministic given $\{z_1, z_2\}$. The state spaces of $X_1$ and $X_2$ are also the set of real numbers.

### 2.2 Mixed Potentials

A *mixed potential* $\zeta$ in a hybrid Bayesian network for a multi-dimensional variable $\mathbf{X} = \mathbf{Y} \cup \mathbf{Z}$ is a pair

$$\zeta(\mathbf{x}) = (\alpha(\mathbf{y}, \mathbf{z}), \phi_\mathbf{y}(\mathbf{z})) \;, \qquad (1)$$

where $\alpha(\mathbf{y}, \mathbf{z})$ is a mass potential and $\phi_\mathbf{y}(\mathbf{z})$ is a density potential. We refer to the first and second parts of the mixed potential as the *mass* and *density* parts, respectively. If the potential represents a deterministic relationship, $\phi_\mathbf{y}(\mathbf{z})$ in (1) is replaced by a linear equation $[g_\mathbf{y}(\mathbf{z}) = 0]$.

### 2.2.1 Mass Potential

Mass potentials are assigned to the mass part of a mixed potential. A *mass potential* $\alpha$ for $\mathbf{X}$ assigns a positive real number to some countable elements in $\Omega_\mathbf{X}$. Since $\mathbf{X}$ has some continuous variables, $\Omega_\mathbf{X}$ has an uncountable number of states and it is only possible to list the values for the countable number of states that have positive mass.

**Example 2** In the Bayesian network in Figure 3, $P(Y_1 = 0) = 0.6$ and $P(Y_1 = 1) = 0.4$, which is defined by the mass potential $\alpha$ where $\alpha(Y_1 = 0) = 0.6$ and $\alpha(Y_1 = 1) = 0.4$. The mixed potential $\eta$ for $Y_1$ has values $\eta(Y_1 = 0) = (0.6, \iota)$ and $\eta(Y_1 = 1) = (0.4, \iota)$.

### 2.2.2 Density Potential

Density potentials can be assigned to the density part of a mixed potential. A *density potential* $\phi$ for $\mathbf{Z}$ is a function $\phi : \Omega_\mathbf{Z} \to \mathcal{R}^+$, where $\mathcal{R}^+$ denotes the set of non-negative real numbers.

**Example 3** In the Bayesian network in Figure 3, $Z_1$ is a standard normal random variable, i.e. $\mathcal{L}(Z_1) \sim N(0,1)$, and the probability distribution for $Z_2$ given $X_1$ is a conditional linear Gaussian distribution, i.e. $\mathcal{L}(Z_2 \mid x_1) \sim N(0.6x_1, 1)$. These distributions are represented by the corresponding density potentials $\phi_1$ for $Z_1$ and $\phi_2$ for $\{X_1, Z_2\}$. The mixed potential $\zeta_1$ for $Z_1$ is $\zeta_1(z_1) = (1, \phi_1(z_1))$. The mixed potential $\zeta_2$ for $\{X_1, Z_2\}$ is $\zeta_2(x_1, z_2) = (1, \phi_2(x_1, z_2))$.

### 2.2.3 Deterministic Potential

A deterministic potential can also be assigned to the density part of a mixed potential. A *deterministic potential* describes the linear deterministic relationship between a set of variables $\mathbf{Z} = \{Z_1, \ldots, Z_n\}$. A deterministic potential for $\mathbf{Z}$ is defined as an equation

$$[g(\mathbf{z}) = 0] = \{w_p \cdot [g_p(\mathbf{z}) = 0]\}_{p=1}^P \, , \qquad (2)$$

where $w_p$, $p = 1, \ldots, P$ are constants. The equation $g_p(\mathbf{z}) = 0$ defines a linear deterministic relationship, where $g_p(\mathbf{z}) = a_{p1}z_1 + \cdots + a_{pn}z_n + b_p$, and where $a_{p1}, \ldots, a_{pn}$ and $b_p$ are real numbers. The coefficient $a_{pj}$ on $Z_j \in \mathbf{Z}$ is equal to 1 when the deterministic potential is specified as a conditional potential for $Z_j$ given $\mathbf{Z}\backslash Z_j$. The *factors* $w_p \cdot [g_p(\mathbf{z}) = 0]$ are maintained as a decomposed set of weighted equations.

**Example 4** In the Bayesian network in Figure 3, $X_1$ is conditionally deterministic given $Y_1$ and $Z_1$. This relationship is represented by the deterministic potential $[g_0(x_1, z_1) = 0]$ or $[x_1 - 2z_1 + 1 = 0]$ if $Y_1 = 0$ and by the deterministic potential $[g_1(x_1, z_1) = 0]$ or $[x_1 - 0.25z_1 - 1 = 0]$ if $Y_1 = 1$. The variable $X_2$ is conditionally deterministic given $\{Z_1, Z_2\}$. This relationship is represented by the deterministic potential $[g_2(x_2, z_1, z_2) = 0]$ or $[x_2 - 0.4z_1 - 0.75z_2 = 0]$. The mixed potential $\xi_1$ for $\{X_1, Y_1, Z_1\}$ has values $\xi_1(x_1, Y_1 = 0, z_1) = (1, [x_1 - 2z_1 + 1 = 0])$ and $\xi_1(x_1, Y_1 = 1, z_1) = (1, [x_1 - 0.25z_1 - 1 = 0])$. The mixed potential $\xi_2$ for $\{X_2, Z_1, Z_2\}$ is $\xi_2(x_2, z_1, z_2) = (1, [x_2 - 0.4z_1 - 0.75z_2 = 0])$.

**Example 5** In the Bayesian network of Example 3, removal of $Y_1$ from the combination of the mixed potential $\xi_1$ for $\{X_1, Y_1, Z_1\}$ and the mixed potential $\eta$ for $Y_1$ results in a mixed potential with the following deterministic potential assigned to the density part: $\{0.6 \cdot [x_1 - 2z_1 + 1 = 0], 0.4 \cdot [x_1 - 0.25z_1 - 1 = 0]\}$.

### 2.2.4 Identity Density Potential

An *identity density potential* $\iota_\mathbf{Z}$ for $\mathbf{Z}$ is a density potential for $\mathbf{Z}$ such that $\iota_\mathbf{Z}(\mathbf{z}) = 1$ for all $\mathbf{z} \in \Omega_\mathbf{Z}$. Further, if $\psi$ is a density or deterministic potential which includes $\mathbf{Z}$ in its domain, $\psi \otimes \iota_\mathbf{Z} = \psi$. For convenience we may drop the subscript and write any identity potential as $\iota$. An identity density potential is used to construct mixed potentials for discrete variables, as shown in Example 2.

## 2.3 Mixtures of Truncated Exponentials (MTE) Density Potentials

A mixture of truncated exponentials (MTE) [Moral *et al.* 2001] density potential has the following definition.

*MTE density potential.* Let $\mathbf{Z} = (Z_1, \ldots, Z_n)$ be an $n$-dimensional random variable. A function $\phi : \Omega_\mathbf{Z} \mapsto \mathcal{R}^+$ is an MTE density potential if one of the next two conditions holds:

1. The potential $\phi$ can be written as

$$\phi(\mathbf{z}) = a_0 + \sum_{i=1}^m a_i \exp\left\{\sum_{j=1}^n b_i^{(j)} z_j\right\} \qquad (3)$$

   for all $\mathbf{z} \in \Omega_\mathbf{Z}$, where $a_i, i = 0, \ldots, m$ and $b_i^{(j)}$, $i = 1, \ldots, m$, $j = 1, \ldots, n$ are real numbers.

2. The domain of the variables, $\Omega_\mathbf{Z}$, is partitioned into hypercubes $\{\Omega_{\mathbf{Z}_1}, \ldots, \Omega_{\mathbf{Z}_k}\}$ such that $\phi$ is defined as $\phi(\mathbf{z}) = \phi_i(\mathbf{z})$ if $\mathbf{z} \in \Omega_{\mathbf{Z}_i}$, $i = 1, \ldots, k$, where each $\phi_i, i = 1, ..., k$ can be written in the form of equation (3) (i.e. each $\phi_i$ is an MTE density potential on $\Omega_{\mathbf{Z}_i}$).

In the definition above, $k$ is the number of *pieces* and $m$ is the number of exponential *terms* in each piece of the MTE density potential. Since the terms in the exponent are a linear function of the arguments, MTE potentials can be easily marginalized in closed form (without having to do computationally expensive numerical integration). And as we shall see later, the class of MTE potentials is closed under combination and marginalization operations.

**Example 6** A 2-piece, 3-term MTE density potential $\phi_1$ which approximates the standard normal PDF is defined [Cobb and Shenoy 2003] as

$$\phi'_1(z_1) = \begin{cases} \begin{array}{l} -0.0105929 \\ +197.5892111 \exp\{2.2568434 z_1\} \\ -462.6885096 \exp\{2.3434117 z_1\} \\ +265.5099139 \exp\{2.4043270 z_1\} \\ \quad \text{if } -3 \leq z_1 < 0 \end{array} \\ \begin{array}{l} -0.0105929 \\ +197.5892111 \exp\{-2.2568434 z_1\} \\ -462.6885096 \exp\{-2.3434117 z_1\} \\ +265.5099139 \exp\{-2.4043270 z_1\} \\ \quad \text{if } 0 \leq z_1 \leq 3 \end{array} \\ 0 \qquad \text{elsewhere.} \end{cases}$$

The MTE density potential $\phi_1$ is assigned as the density part of the mixed potential $\zeta_1$ for $Z_1$ in the Bayesian network of Figure 3. The density potential $\phi_2$ for $\{X_1, Z_2\}$ is defined according to the MTE approximation to the normal PDF in Cobb and Shenoy [2003] with $\mu = 0.6 x_1$ and $\sigma^2 = 1$. The MTE density potential $\phi_2$ is assigned as the density part of the mixed potential $\zeta_2$ for $\{X_1, Z_2\}$ in the Bayesian network of Figure 3.

## 3 OPERATIONS WITH MIXED POTENTIALS

This section describes operations used to propagate potentials of the type described in Section 2.2 in hybrid Bayesian networks using the Shenoy-Shafer architecture. Most definitions are preceded by an example that illustrates the associated operations.

### 3.1 Combination

**Example 7** Let $\zeta_1 = (\alpha_1, \phi_1)$ be a mixed potential for $\{Y_1, Z_1, Z_2\}$ and let $\zeta_2 = (\alpha_2, \phi_2)$ be a mixed potential for $\{Y_1, Y_2, Z_1\}$, where $\alpha_1$ and $\alpha_2$ are mass potentials for $\{Y_1\}$ and $\{Y_1, Y_2\}$, respectively, and $\phi_1$ and $\phi_2$ are density potentials for $\{Z_1, Z_2\}$ and $\{Z_1\}$, respectively. The combination of $\zeta_1$ and $\zeta_2$, denoted by $\zeta = \zeta_1 \otimes \zeta_2$, is the mixed potential $\zeta = (\{\alpha_1, \alpha_2\}, \{\phi_1, \phi_2\})$ for $\{Y_1, Y_2, Z_1, Z_2\}$.

**Definition 1** *Combination of mixed potentials is LAZY combination of the mass and density parts.*

According to the LAZY propagation scheme [Madsen and Jensen 1999] the parts of the mixed potentials are not combined, but rather maintained as a decomposed set of potentials during combination. The usual definition of combination—to consolidate the information in multiple density or mass potentials via pointwise multiplication—is only invoked when marginalization is required.

### 3.2 Marginalization

#### 3.2.1 Discrete Variables

Marginalization of a discrete variable from a mixed potential is illustrated using the following example.

**Example 8** Let $\zeta = (\{\alpha, \beta\}, \phi)$ be a mixed potential for $\{Y_1, Y_2, Z_1\}$ where $\alpha$ is a mass potential for $\{Y_1\}$ with values $\alpha(Y_1 = 0) = 0.7$ and $\alpha(Y_1 = 1) = 0.3$, and $\beta$ is a mass potential for $\{Y_1, Y_2\}$ with values $\beta(Y_1 = 0, Y_2 = 0) = 0.6$, $\beta(Y_1 = 0, Y_2 = 1) = 0.4$, $\beta(Y_1 = 1, Y_2 = 0) = 0.2$, and $\beta(Y_1 = 1, Y_2 = 1) = 0.8$. Let $\phi$ be a density potential for $\{Y_1, Y_2, Z_1\}$ with values $\phi_{i,j}(z_1)$ given $Y_1 = i$ for $i = 0, 1$ and $Y_2 = j$ for $j = 0, 1$. The marginal of $\zeta$ for $\{Y_2, Z_1\}$, denoted by $\zeta'$ or by $\zeta^{\downarrow\{Y_2, Z_1\}}$ or by $\zeta^{-Y_1}$, is computed as

$$\zeta'(y_2, z_1) = \left(1, \{(\alpha \otimes \beta)(y_1, y_2) \cdot \phi_{i,j}(z_1)\}_{\Omega_{Y_1}}\right)$$

for all $(y_2, z_1) \in \Omega_{\{Y_2, Z_1\}}$. Specifically, values for $\zeta'$ are

$$\zeta'(Y_2 = 0, z_1) = (1, \{0.42 \cdot \phi_{0,0}(z_1), 0.06 \cdot \phi_{1,0}(z_1)\}) \ ,$$
$$\zeta'(Y_2 = 1, z_1) = (1, \{0.28 \cdot \phi_{0,1}(z_1), 0.24 \cdot \phi_{1,1}(z_1)\}) \ .$$

**Definition 2** *Marginalization of a discrete variable from a mixed potential is pointwise multiplication of the mass potentials, followed by weighting of the density potentials with the result of the combination. The mass part of the potential is transferred to the density part.*

Although the density parts of the mixed potentials in Example 8 and Definition 2 contain density potentials, the density parts could contain any combination of density potentials and deterministic potentials and the definition holds. In Definition 2, the density values could be weighted with conditional probabilities, but we use the joint probabilities to avoid divisions and normalize after calculating marginals for all variables in the network. By avoiding divisions, we ensure that all resulting density functions remain in the class of MTE potentials.

### 3.2.2 Continuous Variables—Density Potentials

The marginalization operation for a continuous variable required when the density part of a mixed potential contains density potentials is illustrated in the following example.

**Example 9** Let $\zeta = (\alpha, \{\phi_1, \phi_2\})$ be a mixed potential for $\{Y_1, Z_1, Z_2\}$ where $\alpha$ is a mass potential for $\{Y_1, Z_1, Z_2\}$ and $\phi_1$ and $\phi_2$ are density potentials for $\{Z_1, Z_2\}$. The marginal of $\zeta$ for $\{Y_1, Z_2\}$ is computed as

$$\zeta^{\downarrow \{Y_1, Z_2\}}(y_1, z_2) = \left(1, (\alpha \otimes \phi_1 \otimes \phi_2)^{-Z_1}\right)(y_1, z_2)$$
$$= \left(1, \int_{\Omega_{Z_1}} \alpha(y_1, z_1, z_2) \cdot \phi_1(z_1, z_2) \cdot \phi_2(z_1, z_2) \, dz_1\right)$$

for all $(y_1, z_2) \in \Omega_{\{Y_1, Z_2\}}$.

**Definition 3 (Case 1—Non-Empty Set of Remaining Continuous Variables)** *When a mixed potential with density potentials in its density part contains two or more continuous variables in its domain, marginalization of a continuous variable from a mixed potential is pointwise multiplication of the mass and density parts followed by integration over the state space of the continuous variable being removed.*

In Definition 3, if the mass potential does not depend on the continuous variable being removed, it does not have to be combined with the density potentials prior to marginalization and can remain in the mass part of the resulting potential.

A second definition for marginalization of a continuous variable from a mixed potential containing density potentials is required when only one continuous variable remains in the domain of the mixed potential. This is illustrated in the following example.

**Example 10** Let $\zeta = (\alpha, \{\phi_1, \phi_2\})$ be a mixed potential for $\{Y_1, Z_1\}$ where $\alpha$ is a mass potential for $\{Y_1, Z_1\}$ and $\phi_1$ and $\phi_2$ are density potentials for $\{Y_1, Z_1\}$. The marginal of $\zeta$ for $Y_1$ is a mass potential computed as

$$\zeta^{\downarrow Y_1}(y_1) = \left((\alpha \otimes \phi_1 \otimes \phi_2)^{-Z_1}, \iota\right)(y_1)$$
$$= \left(\int_{\Omega_{Z_1}} \alpha(y_1, z_1) \cdot \phi_{1, y_1}(z_1) \cdot \phi_{2, y_1}(z_1) \, dz_1, \iota\right)$$

for all $y_1 \in \Omega_{Y_1}$.

**Definition 4 (Case 2—Empty Set of Remaining Continuous Variables)** *When a mixed potential containing density potentials in its density part contains one continuous variable in its domain, marginalization of the continuous variable from the mixed potential is pointwise multiplication of the mass and density parts followed by integration over the state space of the continuous variable. The result of the integration is a constant which becomes the mass part of the potential. The density part of the potential is replaced by an identity density potential.*

### 3.2.3 Continuous Variables—Deterministic Potentials

The marginalization operation for continuous variables required when the density part of a mixed potential contains deterministic potentials is illustrated in the following example.

**Example 11** Suppose two deterministic potentials are defined as:

$$\{0.7 \cdot [-2z_1 + z_2 - 1 = 0], 0.3 \cdot [3z_1 + z_2 - 2 = 0]\},$$
$$\{0.1 \cdot [-3z_1 - 2z_2 + z_3 - 1 = 0],$$
$$0.9 \cdot [3z_1 - 2z_2 + z_3 + 2 = 0]\} \, .$$

The marginal deterministic potential created by removing $Z_2$ from the combination of these two deterministic potentials is computed as

$$\{0.07 \cdot [-7z_1 + z_3 - 3 = 0], 0.63 \cdot [-z_1 + z_3 = 0],$$
$$0.03 \cdot [-11z_1 + z_3 - 3 = 0], 0.27 \cdot [-3z_1 + z_3 - 2 = 0]\}.$$

**Definition 5** *Marginalization of a continuous variable $Z_i$ from a mixed potential with two deterministic potentials in its density part is substitution of the inverse of one equation with respect to $Z_i$ into the second equation, along with multiplication of the weights on the deterministic potentials.*

If the deterministic potentials contains multiple factors, the definition is applied separately to each pairwise combination of factors, as demonstrated in Example 11. An extension of Definition 5 required for the case where a mixed potential contains three or more deterministic potentials is described in [Cobb and Shenoy 2005a].

An additional marginalization operation is required when a mixed potential contains a single deterministic potential in its density part, as stated in the following definition.

**Definition 6** *Marginalization of a continuous variable from a single deterministic potential results in an identity density potential for the remaining variables.*

The operation described in Definition 6 is motivated as follows. A deterministic potential for a set of variables can be regarded as a conditional for any one variable given the others. Thus when we marginalize a variable from a deterministic potential, the resulting marginal is vacuous.

### 3.2.4 Continuous Variables—Density and Deterministic Potentials

After combination, a mixed potential may contain a density potential and a deterministic potential in its density part. The following example illustrates the marginalization operation required for continuous variables in this case.

**Example 12** Suppose $\phi$ is a density potential for $\{Z_1, Z_3\}$ and a deterministic potential for $\{Z_1, Z_3, Z_4\}$ is defined as

$$\{0.6 \cdot [-0.5z_1 + 0.25z_3 + z_4 = 0], 0.4 \cdot [-3z_1 + 2z_3 + z_4 = 0]\} \ .$$

The marginal density potential created by removing $Z_1$ from the combination of these potentials is

$$\phi'(z_3, z_4) = \left\{ 0.6 \cdot \frac{1}{|-0.5|} \cdot \phi\left(\frac{-0.25z_3 - z_4}{-0.5}, z_3\right), \right.$$
$$\left. 0.4 \cdot \frac{1}{|-3|} \cdot \phi\left(\frac{-2z_3 - z_4}{-3}, z_3\right) \right\} \ .$$

**Definition 7** *Marginalization of a continuous variable $Z_i$ from the combination of a density potential and a deterministic potential (defined where $Z_i$ is in the domain of both potentials and the equation(s) in the deterministic potential are invertible in $Z_i$) is substitution of the inverse of the equation(s) in the deterministic potential into the product of the density potential and the mass potential, with the result multiplied by $1/|a_i|$, where $a_i$ is the coefficient on $Z_i$ in the equation(s) of the deterministic potential.*

This operation is derived from the method of convolutions in probability theory, as shown in [Cobb and Shenoy 2004a]. Mao *et al.* [2004] also discusses convolution operations in the context of probabilistic models. An extension of Definition 7 required for the case where a mixed potential contains a density potential and multiple deterministic potentials is described in [Cobb and Shenoy 2005a].

## 3.3 Restriction

Restriction—or entering evidence—is a marginalization operation which involves dropping coordinates to define a potential on a smaller set of variables.

### 3.3.1 Discrete Variables

Restriction of a mixed potential to an observed state for a discrete variable is illustrated in the following example.

**Example 13** Suppose a mixed potential is defined for $\{Y_1, Z_1, Z_2\}$ as $\zeta(y, z_1, z_2) = (0.6, \phi_y(z_1, z_2))$ and $\zeta(ny, z_1, z_2) = (0.4, \phi_{ny}(z_1, z_2))$.

A potential $\xi(Y_1 = ny) = (1, \iota)$ for $Y_1$ characterizes the evidence $Y_1 = ny$. The restriction of $\zeta$ to $Y_1 = ny$, denoted by $\zeta' = (\zeta \otimes \xi)^{-Y_1}$, results in a potential $\zeta'$ for $\{Z_1, Z_2\}$ defined as $\zeta'(z_1, z_2) = (0.4, \phi_{ny}(z_1, z_2))$.

**Definition 8** *Restriction of a discrete variable to an observed state that has positive mass leads to zeroing out all masses for unobserved states.*

### 3.3.2 Continuous Variables—States with Positive Mass

Continuous variables are those whose state spaces are non-countable. A continuous variable may have states with positive mass (as shown in the example of Figure 2). The following example illustrates the restriction operation used when observing a state of a continuous variable with positive mass.

**Example 14** Suppose a mixed potential is defined for $\{X, Y, Z\}$ as $\zeta(X = 1, Y = 1, z) = (0.5, \iota)$, $\zeta(x, Y = 2, z) = (1, [x - z = 0])$, and $\zeta(x, Y = 3, z) = (1, \phi(x, z))$. A potential $\xi(X = 1) = (1, \iota)$ for $X$ characterizes the evidence $X = 1$. The restriction of $\zeta$ to $X = 1$, denoted by $\zeta' = (\zeta \otimes \xi)^{-X}$, results in a potential $\zeta'$ for $\{Y, Z\}$ defined as $\zeta'(Y = 1, z) = (0.5, \iota)$.

**Definition 9** *Restriction of a continuous variable to an observed state that has a mass leads to zeroing out all masses and densities for unobserved states.*

### 3.3.3 Continuous Variables—Density Potentials

The following restriction operations are required when a variable whose state space is non-countable is restricted to a value that has a non-vacuous density.

**Example 15** Suppose a mixed potential is defined for $\{Y_1, Z_1, Z_2\}$ as $\zeta(y, z_1, z_2) = (0.6, \phi_y(z_1, z_2))$ and $\zeta(ny, z_1, z_2) = (0.4, \phi_{ny}(z_1, z_2))$. A potential $\xi(Z_2 = 5) = (1, \iota)$ for $Z_2$ characterizes the evidence $Z_2 = 5$. The restriction of $\zeta$ to $Z_2 = 5$, denoted by $\zeta' = (\zeta \otimes \xi)^{-Z_2}$, results in a potential $\zeta'$ for $\{Y_1, Z_1\}$ defined as $\zeta'(y, z_1) = (0.6, \phi_y(z_1, 5))$ and $\zeta'(ny, z_1) = (0.4, \phi_{ny}(z_1, 5))$.

**Definition 10** (**Case 1–Non-Empty Set of Remaining Continuous Variables**) *When a mixed potential has a domain with two or more continuous variables, the restriction of a continuous variable to a value* with a non-vacuous density is substitution of a real number into the density potential, resulting in a density potential for the remaining variables.

An additional restriction operation is required for the case where the observed variable is the only continuous variable in the domain of the mixed potential, as illustrated in the following example.

**Example 16** Suppose a mixed potential is defined for $\{Y_1, Z_1\}$ as $\zeta(y, z_1) = (0.2, \phi_y(z_1))$ and $\zeta(ny, z_1) = (0.8, \phi_{ny}(z_1))$. A potential $\xi(Z_1 = 0) = (1, \iota)$ for $Z_1$ characterizes the evidence $Z_1 = 0$. The restriction of $\zeta$ to $Z_1 = 0$, denoted by $\zeta' = (\zeta \otimes \xi)^{-Z_1}$, results in a mass potential $\zeta'$ for $Y_1$: $\zeta'(y) = (0.2 \cdot \phi_y(0), \iota)$ and $\zeta'(ny) = (0.8 \cdot \phi_{ny}(0), \iota)$.

**Definition 11** (**Case 2–Empty Set of Remaining Continuous Variables**) *When a mixed potential has a domain with one continuous variable, the restriction of the continuous variable to a value with a non-vacuous density potential is substitution of a real number into the density potential, multiplication of the result by the mass part of the potential, and replacement of the density part of the potential with an identity density potential.*

### 3.3.4 Continuous Variables—Deterministic Potentials

The following examples illustrate different cases of restriction operations on deterministic potentials.

**Example 17** Suppose a potential $\zeta_1$ is defined for $\{Y_1, Z_1, Z_2, Z_3\}$ as follows:

$\zeta_1(y, z_1, z_2, z_3) = (1, \{0.2 \cdot [-2z_1 - 0.75z_2 + z_3 = 0],$
$0.8 \cdot [-3z_1 - z_2 + z_3 = 0]\})$
$\zeta_1(ny, z_1, z_2, z_3) = (1, \{0.9 \cdot [-5z_1 - 0.2z_2 + z_3 = 0],$
$0.1 \cdot [-0.4z_1 - 0.1z_2 + z_3 = 0]\})$ .

The evidence $Z_1 = 2$ is characterized by the potential $\xi(Z_1 = 2) = (1, \iota)$ for $Z_1$. The restricted potential $\zeta_1'$ for $\{Y_1, Z_2, Z_3\}$ created by removing $Z_2$, an operation denoted $\zeta_1' = (\zeta_1 \otimes \xi)^{-Z_1}$, is defined as

$\zeta_1'(y, z_2, z_3) = (1, \{0.2 \cdot [-0.75z_2 + z_3 - 4 = 0],$
$0.8 \cdot [-z_2 + z_3 - 6 = 0]\})$
$\zeta_1'(ny, z_2, z_3) = (1, \{0.9 \cdot [-0.2z_2 + z_3 - 10 = 0],$
$0.1 \cdot [-0.1z_2 + z_3 - 0.8 = 0]\})$ .

**Definition 12** (**Case 1–Multiple Remaining Continuous Variables**) *When a mixed potential with deterministic potentials in its density part has a domain containing three or more continuous variables,* restriction of a continuous variable to a value with non-vacuous density is substitution of a real number into the equation(s) in the deterministic potential.

If the deterministic potential contains multiple factors, the operation in Definition 12 is applied separately to each factor, as demonstrated in Example 17.

An additional operation for deterministic potentials is required when a mixed potential has two continuous variables in its domain and one variable is restricted to an observed value with non-vacuous density. This is shown in the following example.

**Example 18** Consider the following potential $\zeta$ defined for $\{Y_1, Z_1, Z_2\}$:

$\zeta(y, z_1, z_2) = (0.6, [2z_1 - 3z_2 + 2 = 0])$
$\zeta(ny, z_1, z_2) = (0.4, [3z_1 + 5z_2 + 2 = 0])$ .

The evidence $Z_2 = 0$ is characterized by the potential $\xi(Z_2 = 0) = (1, \iota)$. The restricted potential $\zeta'$ for $\{Y_1, Z_1\}$ created by removing $Z_1$ is $\zeta' = (\zeta \otimes \xi)^{-Z_2}$, defined as $\zeta'(y, Z_1 = -1) = (0.6 \cdot 1/2, \iota)$ and $\zeta'(ny, Z_1 = -2/3) = (0.4 \cdot 1/3, \iota)$. The mass parts of the potential are multiplied by likelihoods since the result of substituting the evidence is a constant. The result of the operation is a mass potential for $\{Y_1, Z_1\}$.

**Definition 13** (**Case 2–One Remaining Continuous Variable**) *When a mixed potential has a domain containing two continuous variables $Z_1$ and $Z_2$ and deterministic variables in its density part, restriction of the deterministic potential to $Z_2 = c$ is substitution of a real number $c$ into the equation(s) in the deterministic potential to obtain a restricted value for $Z_1$. The mass part of the potential is multiplied by the existing weight in the deterministic potential times a likelihood equal to $1/|a_1|$, where $a_1$ is the coefficient on $Z_1$ in the equation(s) of the deterministic potential.*

The need for the likelihood $1/|a_1|$ in Definition 13 can be explained as follows. If we marginalize $Z_1$ by combining $[g_{\mathbf{y}}(z_1, z_2) = 0]$ with a density potential $\phi_{\mathbf{y}}$ for $Z_1$, the normalization constant $1/|a_1|$ is required according to Definition 7. Thus, the same normalization constant is required if we marginalize $Z_1$ after entering evidence on $Z_2$, so it must be included as a likelihood when creating the mass potential in Definition 13.

If a mixed potential contains density potentials and deterministic potentials, the appropriate restriction definition is applied to each type of potential if a continuous variable in the domain of each type of potential is restricted to an observed value. Cobb and Shenoy [2005a] demonstrate that the operations used for propagation satisfy the Shenoy-Shafer axioms [Shenoy and Shafer 1990] for local propagation and maintain density potentials that are within the class of MTE density potentials. We can use any join tree for propagation with no restrictions placed on the initialization phase. This is in contrast to the architecture of Lauritzen and Jensen [2001], which requires a strong junction tree so that continuous variables are marginalized before discrete ones.

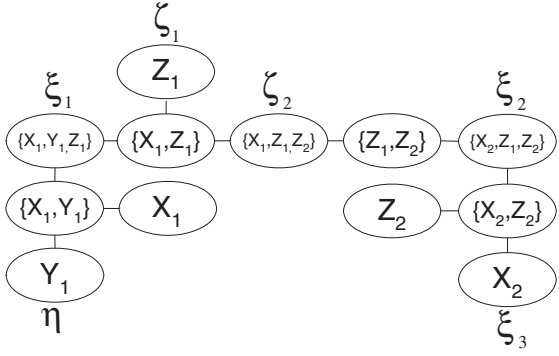

Figure 4: Binary Join Tree for Example Two

## 4 AN EXAMPLE

Consider the Bayesian network in Figure 3 with mixed potentials as defined in the examples of Section 2. Assume evidence that $X_2 = 1$ is characterized by the potential $\xi_3(X_2 = 1) = (1, \iota)$. The binary join tree used to solve the example is shown in Figure 4.

### 4.1 Messages

The relevant messages required to calculate marginals for all variables considering the evidence are detailed below. For convenience, identity potentials are omitted from the details of the calculations.

$\{X_2, Z_1, Z_2\}$ to $\{Z_1, Z_2\}$

This message is determined by restriction as $\xi_4(z_1, z_2) = (\xi_2 \otimes \xi_3)^{-X_2}(z_1, z_2) = (1, [-0.4z_1 - 0.75z_2 + 1 = 0])$.

$\{X_1, Z_1, Z_2\}$ to $\{X_1, Z_1\}$

This message is determined as $\zeta_3(x_1, z_1) = (\zeta_2 \otimes \xi_4)^{-Z_2}(x_1, z_1) = (1, \phi_3(x_1, z_1))$, where $\phi_3(x_1, z_1) = (1/0.75) \cdot \phi_2(x_1, (1 - 0.4z_1)/0.75)$.

$\{X_1, Y_1, Z_1\}$ to $\{X_1, Y_1\}$

This message is determined as $\zeta_4 = (\zeta_1 \otimes \zeta_3 \otimes \xi_1)^{-Z_1}$. The resulting mixed potential $\zeta_4$ for $\{X_1, Y_1\}$ is denoted as $\zeta_4(x_1, Y_1 = 0) = (1, \phi_{4,0}(x_1))$ and $\zeta_4(x_1, Y_1 = 1) = (1, \phi_{4,1}(x_1))$, where $\phi_{4,0}(x_1) = (1/2) \cdot (\phi_1 \otimes \phi_3)(x_1, (x_1 + 1)/2)$ and $\phi_{4,1}(x_1) = (1/0.25) \cdot (\phi_1 \otimes \phi_3)(x_1, (x_1 - 1)/0.25)$.

$\{X_1, Y_1\}$ to $\{Y_1\}$

This message is determined as $\chi = (\zeta_4)^{-X_1}$, which results in mass potentials calculated as $\chi(Y_1 = 0) = \int_{\Omega_{X_1}} \phi_{4,0}(x_1) \, dx_1$ and $\chi(Y_1 = 1) = \int_{\Omega_{X_1}} \phi_{4,1}(x_1) \, dx_1$.

$\{X_1, Y_1\}$ to $\{X_1\}$

This message is determined as $\zeta_5(x_1) = (\eta \otimes \zeta_4)^{-Y_1}(x_1) = (1, \{0.6 \cdot \phi_{4,0}(x_1), 0.4 \cdot \phi_{4,1}(x_1)\})$.

$\{X_1, Y_1, Z_1\}$ to $\{X_1, Z_1\}$

This message is determined as $\xi_5(x_1, z_1) = (\eta \otimes \xi_1)^{-Y_1}(x_1, z_1) = (1, \{0.6 \cdot [x_1 - 2z_1 + 1 = 0], 0.4 \cdot [x_1 - 0.25z_1 - 1 = 0]\})$.

$\{X_1, Z_1\}$ to $\{Z_1\}$

This message is determined as $\zeta_6(z_1) = (\zeta_3 \otimes \xi_5)^{-X_1}(z_1) = (1, \{0.6 \cdot \phi_{5,0}(z_1), 0.4 \cdot \phi_{5,1}(z_1)\})$, where $\phi_{5,0}(z_1) = \phi_3(2z_1 - 1, z_1)$ and $\phi_{5,1}(z_1) = \phi_3(0.25z_1 + 1, z_1)$.

$\{X_1, Z_1, Z_2\}$ to $\{Z_1, Z_2\}$

This message is determined as $\zeta_7(z_1, z_2) = (\zeta_1 \otimes \zeta_2 \otimes \xi_5)^{-X_1}(z_1, z_2) = (\zeta_1 \otimes (\zeta_2 \otimes \xi_5))^{-X_1}(z_1, z_2) = (1, \{0.6 \cdot \phi_{6,0}(z_1, z_2), 0.4 \cdot \phi_{6,1}(z_1, z_2)\})$, where $\phi_{6,0}(z_1, z_2) = \phi_2(2z_1 - 1, z_2)$ and $\phi_{6,1}(z_1, z_2) = \phi_2(0.25z_1 + 1, z_2)$. The message is sent as $\{\zeta_7, \zeta_1\}$.

$\{X_2, Z_1, Z_2\}$ to $\{X_2, Z_2\}$

This message is determined as $\zeta_8(x_2, z_2) = (\zeta_1 \otimes \zeta_7 \otimes \xi_2)^{-Z_1}(x_2, z_2) = (1, \{0.6 \cdot \phi_{7,0}(x_2, z_2), 0.4 \cdot \phi_{7,1}(x_2, z_2)\})$, where $\phi_{7,0}(x_2, z_2) = (1/0.4) \cdot (\phi_1 \otimes \phi_{6,0})((x_2 - 0.75)/0.4, z_2)$ and $\phi_{7,1}(x_2, z_2) = (1/0.4) \cdot (\phi_1 \otimes \phi_{6,1})((x_2 - 0.75)/0.4, z_2)$.

$\{X_2, Z_2\}$ to $\{Z_2\}$

This message is determined as $\zeta_9 = (\zeta_8 \otimes \xi_3)$. After the restriction, the resulting density potentials are $\phi_{8,0}(z_2) = \phi_{7,0}(1, z_2)$ and $\phi_{8,1}(z_2) = \phi_{7,1}(1, z_2)$. The resulting mixed potential is $\zeta_9(z_2) = (1, \{0.6 \cdot \phi_{8,0}(z_2), 0.4 \cdot \phi_{8,1}(z_2)\})$.

### 4.2 Posterior Marginals

Posterior probabilities for $Y_1$ are $\alpha'(Y_1 = 0) = (0.6 \cdot \chi(Y_1 = 0))/K_0$ and $\alpha'(Y_1 = 1) = (0.4 \cdot \chi(Y_1 = 1))/K_0$. The marginal density potential for $Z_1$ is

$\phi_9(z_1) = K_1 \cdot \phi_1(z_1) \cdot (0.6 \cdot \phi_{5,0}(z_1) + 0.4 \cdot \phi_{5,1}(z_1))$. The marginal density potential for $Z_2$ is $\phi_{10}(z_2) = K_2 \cdot (0.6 \cdot \phi_{8,0}(z_2) + 0.4 \cdot \phi_{8,1}(z_2))$. The marginal density potential for $X_1$ is $\phi_{11}(x_1) = K_3 \cdot (0.6 \cdot \phi_{4,0}(x_1) + \phi_{4,1}(x_1))$. $K_i, i = 0, \ldots, 3$ are normalization constants. A comparison of the expected values and variances of the marginal distributions with those calculated in Hugin for several evidence scenarios is given in [Cobb and Shenoy 2005a].

## 5 CONCLUSIONS

The main contribution of this paper is to extend exact inference in hybrid Bayesian networks in which continuous variables may have any conditional density functions (not necessarily conditional linear Gaussian distributions), discrete variables may have continuous parents, and we may have conditionally deterministic continuous variables that are linearly dependent on their continuous parents. The scheme uses a mixed distribution representation of potentials and derives operations from the method of convolutions in probability theory to determine distributions for linear functions of random variables. MTE potentials are used to approximate probability density functions in the representation so that probability density functions can be easily marginalized (without resorting to computationally expensive numerical integration). The Shenoy-Shafer architecture with LAZY propagation is used to calculate marginals. One advantage of our approach is that we can marginalize variables in any order. This is in contrast with the Lauritzen and Jensen [2001] algorithm in which continuous variables have to marginalized before discrete ones. This restriction can lead to very large cliques.

Two practical limitations of our algorithm are as follows. As in Lauritzen and Jensen [2001], the conditionally deterministic variables must be described by a *linear* function of its continuous parents and the conditional probability density functions for continuous variables must be approximated by MTE potentials. Cobb and Shenoy [2005b] show how the first limitation can be overcome by approximating nonlinear functions by piecewise linear functions. Cobb *et al.* [2004b] describes an algorithm for approximating any probability density function by MTE potentials.

**Acknowledgements**

This research was completed while Barry Cobb was a doctoral student at the University of Kansas and supported by a dissertation fellowship from the School of Business. The authors thank the UAI referees for their constructive suggestions which improved the paper.